\documentclass[pdflatex,sn-mathphys-num,normalem]{sn-jnl}
\usepackage{graphicx}
\usepackage{multirow}
\usepackage{amsmath,amssymb,amsfonts}
\usepackage{amsthm}
\usepackage{mathrsfs}
\usepackage[title]{appendix}
\usepackage{xcolor}
\usepackage{textcomp}
\usepackage{manyfoot}
\usepackage{booktabs}
\usepackage{algorithm}
\usepackage{algorithmicx}
\usepackage{algpseudocode}
\usepackage{listings}
\usepackage{textcomp}
\usepackage{booktabs}
\usepackage{hyperref}
\usepackage[figuresright]{rotating}
\usepackage{listings}
\usepackage{cleveref}
\usepackage[normalem]{ulem}
\usepackage{xspace}
\usepackage[version=4]{mhchem}
\usepackage{subcaption}
\usepackage{array}
\usepackage{float}


\usepackage{alphalph}
\usepackage{fnpct}

\Crefname{table}{\text{Table}}{\text{Tables}}
\Crefname{figure}{\text{Fig.}}{\text{Figs.}}
\Crefname{section}{\text{Sec}}{\text{Sections}}
\Crefname{equation}{\text{Eq.}}{\text{Eqs.}}

\usepackage{booktabs}
\usepackage{longtable}
\usepackage{array}
\usepackage{ragged2e}
\usepackage{pdflscape}

\newcolumntype{L}[1]{>{\RaggedRight\arraybackslash}p{#1}}
\newcolumntype{R}[1]{>{\RaggedLeft\arraybackslash}p{#1}}

\begin{document}

\title[Article Title]{Highly accurate population-scale hepato-pancreato-biliary cancer screening with self-evolving trustworthy AI}

\author*[1,2,3]{\fnm{Haojun} \sur{Yu}}\email{haojunyu@pku.edu.cn}\equalcont{These authors contributed equally to this work.}
\author[4]{\fnm{Jiachen} \sur{Zheng}}\nomail\equalcont{These authors contributed equally to this work.}
\author[2]{\fnm{Yingda} \sur{Xia}}\nomail\equalcont{These authors contributed equally to this work.}
\author[5]{\fnm{Haiyan} \sur{Xin}}\nomail\equalcont{These authors contributed equally to this work.}
\author[4]{\fnm{Xiaowen} \sur{Tang}}\nomail
\author[7]{\fnm{Caichao} \sur{Lv}}\nomail 
\author[4]{\fnm{Yiang} \sur{Gao}}\nomail
\author[4]{\fnm{Xiangran} \sur{Jia}}\nomail
\author[4]{\fnm{Wenqing} \sur{Yang}}\nomail
\author[4]{\fnm{Runxin} \sur{Shi}}\nomail
\author[4]{\fnm{Xinhe} \sur{Zhang}}\nomail
\author[7]{\fnm{Ze} \sur{Zhang}}\nomail 
\author[13]{\fnm{Wei} \sur{Qiu}}\nomail 
\author[11]{\fnm{Binbin} \sur{Su}}\nomail 
\author[18]{\fnm{Hui} \sur{Ding}}\nomail 
\author[15]{\fnm{Quanlin} \sur{Wu}}\nomail
\author[5]{\fnm{Yinbo} \sur{Liu}}\nomail
\author[12]{\fnm{Jie} \sur{Xie}}\nomail 
\author[16]{\fnm{Yifeng} \sur{Lan}}\nomail 
\author[1,19]{\fnm{Ashok} \sur{Saluja}}\nomail 
\author[17]{\fnm{Jingshan} \sur{Pan}}\nomail 
\author*[14]{\fnm{Zhike} \sur{Liu}}\email{liu-zhike@bjmu.edu.cn}
\author*[8]{\fnm{Guangyi} \sur{Fan}}\email{fanguangyi@genomics.cn}
\author*[9,10]{\fnm{James} \sur{Zou}}\email{jamesz@stanford.edu} 
\author*[3,15]{\fnm{Liwei} \sur{Wang}}\email{wanglw@pku.edu.cn} 
\author*[7]{\fnm{Jie} \sur{Nie}}\email{niejie@ouc.edu.cn} 
\author*[6]{\fnm{Ning} \sur{Mao}}\email{maoning@pku.edu.cn} 
\author*[1,4]{\fnm{He} \sur{Ren}}\email{herenrh@163.com}

\affil[1]{\orgname{Shandong Provincial Key Laboratory of Clinical Research for Pancreatic Diseases, Tumor Immunology and Cytotherapy, Medical Research Center, The Affiliated Hospital of Qingdao University, Qingdao University, Qingdao, China}}%
\affil[2]{\orgname{Yiyong Technology Group Co., Ltd}}
\affil[3]{\orgname{State Key Laboratory of General Artificial Intelligence, Peking University, Beijing, China}}%
\affil[4]{\orgname{Qingdao Medical College, Qingdao University, Qingdao, China}}
\affil[5]{\orgname{Department of AI and Big Data Management, The Affiliated Hospital of Qingdao University, Qingdao, China}}
\affil[6]{\orgname{Department of Radiology, Yantai Yuhuangding Hospital, Qingdao University, Yantai, China}}% 
\affil[7]{\orgname{The College of Information Science and Engineering, Ocean University of China, Qingdao, China}}%
\affil[8]{\orgname{BGI Research, Qingdao, China}}%
\affil[9]{\orgname{Department of Computer Sciences, Stanford University, Stanford, California, United States}} % 
\affil[10]{\orgname{Department of Biomedical Data Science, Stanford University, Stanford, California, United States}}
\affil[11]{\orgname{School of Population Medicine and Public Health, Peking Union Medical College, Beijing, China}}%
\affil[12]{\orgname{The Laboratory for Economic Behaviors and Policy Simulation, Nankai University, Tianjin, China}}
\affil[13]{\orgname{Department of Electrical and Computer Engineering, Rice University, Texas, United States}}
\affil[14]{\orgname{Peking University-Yinzhou Regional joint institute for Advanced Medical Intelligence and Digital health, Ningbo, China}}
\affil[15]{\orgname{Center for Data Science, Peking University, Beijing, China}}
\affil[16]{\orgname{School of Mathematics and Statistics, Qingdao University, Qingdao, China}}
\affil[17]{\orgname{Key Laboratory of Computing Power Network and Information Security, Qilu University of Technology, Jinan, China}}
\affil[18]{\orgname{Radiology Department, the Affiliated Hospital of Qingdao University, Qingdao University, Qingdao, China}}
\affil[19]{\orgname{Pancreas Center, Center for Interventional Endoscopy, AdventHealth Orlando, Orlando, Florida}}
% %%%%%%%% HOPE %%%%%%%%
\abstract{
Early cancer detection can reduce cancer deaths through radical resection. However, the current implementation of recommended screening may not cover as much as 87.2\% of cancer deaths, because population-scale screening is challenging due to the low cancer prevalence, the high economic cost, and the potential harms of false positives. In this paper, we propose an AI-based risk stratification funnel to progressively concentrate screening efforts on higher-risk candidates, which radically suppresses false positives and enables effective population-scale hepato-pancreato-biliary (HPB) cancer screening. Another challenge is that AI models' clinical utility depends not only on their performance but also on physicians' trust, as they are not authorized to make clinical decisions. To improve the generalizability and interpretability of AI, we design a self-evolving algorithm to incentivize models' reasoning capability. Accordingly, we developed this {H}epat{O}-{P}ancreato-biliary cancer screening approach with self-{E}volving trustworthy AI ({\bf HOPE}) as a three-stage funnel utilizing routine clinical indicators, electronic health records (EHR) and non-contrast computed tomography scans. We developed HOPE on a large-scale internal cohort (n=10,269,153) and conducted extensive evaluations to assess its ability. We demonstrate that HOPE largely reduces false positives, achieving an extrapolated positive predictive value of 80.7\% and substantially outperforming state-of-the-art (SOTA) single-test approaches (10\%-49\%). Moreover, using only EHR data, our AI model is capable of predicting high-risk candidates before standard of care with an average lead time of 1.53-2.94 years. Also, the reasoning capability significantly improves the generalizability of AI on external cohorts (United Kingdom n=501,946; United States n=364,627; Real-world n=139,373) compared to baselines. Importantly, the reasoning processes foster physician trust and improve the mean acceptance rate from 16.6\% to 64.1\%. Finally, we indicate that HOPE is cost-effective compared with SOTA single-test approaches. In sum, HOPE could serve as the first effective approach for population-scale HPB cancer screening.
}

\maketitle

\section{Main}
\label{sec:main}

Cancer is a leading cause of death worldwide, accounting for approximately 10 million deaths annually~\cite{bray2024global}. An important approach to reducing cancer deaths is screening~\cite{nelson2016effectiveness,lin2016screening,national2011reduced}, as the majority of early-stage cancers can be cured by radical resection~\cite{siegel2025cancer,etzioni2003case}. Screening of these cancers poses challenges for physicians due to low prevalence and potential harms of false positives~\cite{etzioni2003case,banales2020cholangiocarcinoma,waller2015hepatocellular,yamada2023advances,el2019intrahepatic,loud2017cancer}. Recommended screening approaches are only available for a few prevalent cancers~\cite{national2011reduced,shapiro1971periodic,myers2015benefits}. In contrast, most cancers lack population-scale screening approaches~\cite{cohen2018detection}. As a result, the current implementation of recommended screening may not cover as much as 87.2\% of cancer deaths~\cite{ofman2025estimated}. {\bf Thus, there exists an urgent need to develop an effective approach for population-scale cancer screening.}
Recently, AI models have made great progress in accurate cancer screening. Previous studies demonstrate that AI models are capable of detecting early-stage cancers using medical images~\cite{wang2021development,cao2023large,
hu2025ai,lotter2021robust,shi2025artificial}, liquid biopsy~\cite{chen2020non,klein2021clinical,wang2025development,metcalf2024micrornas} or electronic health records (EHR)~\cite{rasmy2021med,placido2023deep,mishra2024machine,park2024enhancing,kraljevic2024foresight,fahrner2025generative}.

Despite the progress, there are at least three challenges~\cite{chen2021artificial,lei2025artificial,lynge2026challenges,ten2021personalising,volk2024american} that hinder the real-world application of population-scale cancer screening. {\bf First}, even with high predictive performance, single-test approaches yield a substantial number of false positives when deployed at the population scale, because of the low cancer prevalence. The extrapolated positive predictive values (PPV) of state-of-the-art (SOTA) single-test approaches are about 10\%-49\%~\cite{liu2020sensitive,cao2023large,nadauld2021pathfinder,schrag2023pathfinder}, implying that the identification of each cancer patient is accompanied by at least one false positive. These false positives can cause unnecessary anxiety and unwarranted interventions, posing a major barrier to real-world application. {\bf Second}, while EHRs are routinely generated and essential for scalable screening, the robust generalizability of AI models on EHR data remains suboptimal~\cite{placido2023deep,liu2026advancing}. Previous studies show that the variations in patient demographics, ethnicity, and clinical documentation practices can lead to substantial performance degradation across external cohorts~\cite{placido2023deep,singhi2019early,xia2023cancer,miotto2018deep}. {\bf Third}, physicians may distrust the SOTA large language models (LLMs) in analyzing patient EHR~\cite{goh2024large,johri2025evaluation,asan2020artificial}, yet AI models are not authorized to make clinical decisions. Therefore, it is important to build interpretable AI models to foster physician trust and ultimately improve their clinical utility. However, the lack of expert consensus on interpreting patient EHRs hinders the formalization of gold-standard reasoning labels for developing AI models.

\begin{figure}[t]
    \centering
    \includegraphics[width=\linewidth]{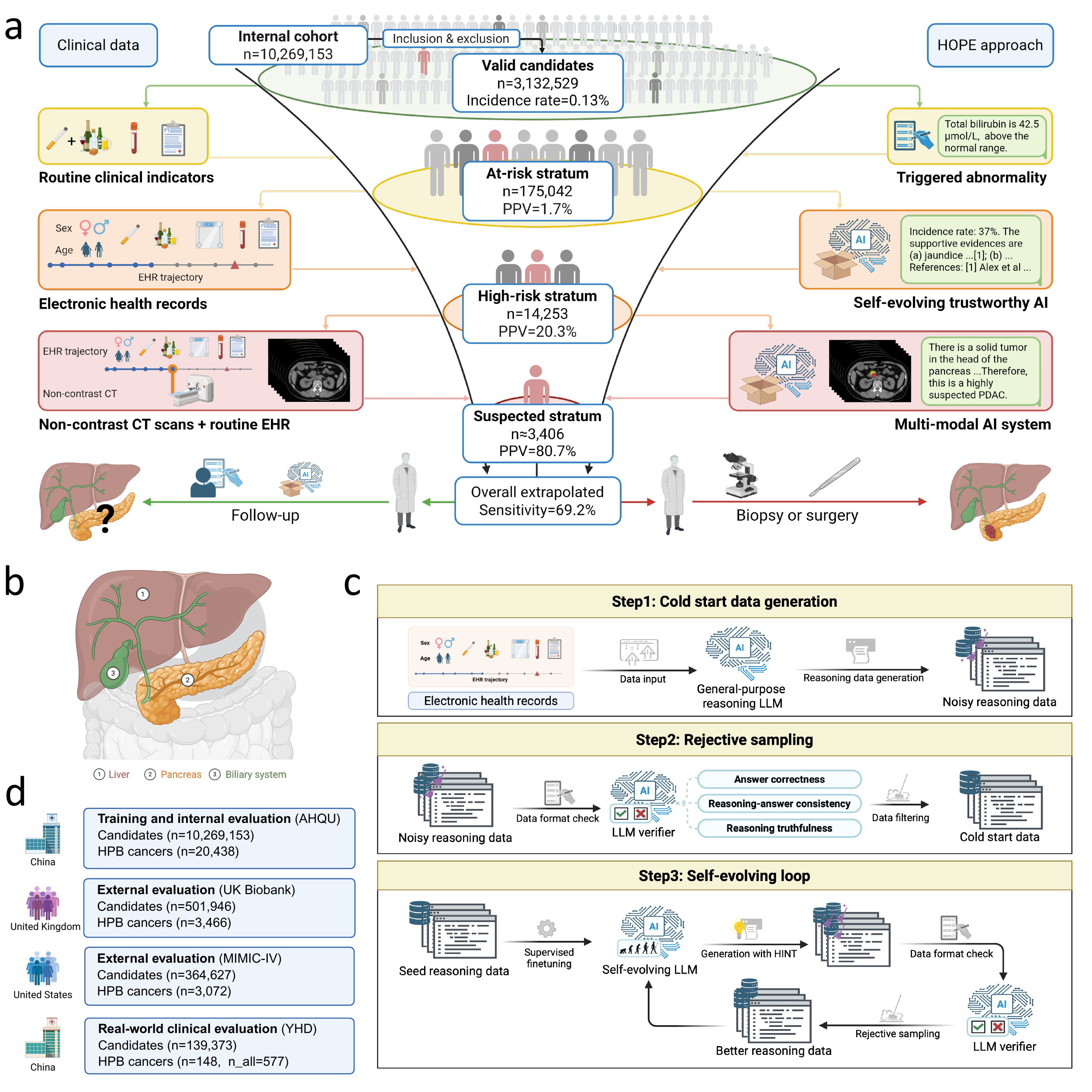}
    \caption{\small \small {\bf The overview of HOPE.} {\bf a}, The three-stage risk stratification screening funnel enables a radical suppression of false positives. The evaluation metrics are in patient-level. The number of suspected stratum candidates (n$\approx$3,406) was extrapolated based on the performance of multi-modal AI system and the incidence rate of high-risk stratum.
    {\bf b}, The hepato-pancreato-biliary (HPB) system. The liver, gallbladder, and pancreas are functionally interrelated digestive organs sharing interconnected biliary and vascular systems.
    {\bf c}, Self-evolving trustworthy AI. This algorithm iteratively improves the reasoning capability of large language models (LLMs) by training on progressively higher-quality data, which are generated by the model in the previous iteration and verified by strict LLM verifiers.
    {\bf d}, Overview of the internal and external cohorts. The real-world cohort comprised a continuous series of patients, among whom 577 were confirmed to have HPB cancer during follow-up, including 148 patients who developed HPB cancer after cohort entry.}
    \label{fig:overview}
\end{figure}

To address these challenges, we propose a new multi-cancer screening approach and apply it to hepato-pancreato-biliary (HPB) cancers, as shown in \Cref{fig:overview}a. HPB cancers are highly lethal diseases that collectively cause more than 1.3 million deaths annually~\cite{bosch2004primary,balogh2016hepatocellular,chan2025lancet,kamisawa2016pancreatic,valle2021biliary,roa2022gallbladder,bray2024global}. {\bf Importantly, there are no recommended screening approaches for HPB cancers.} The 5-year survival rate for HPB cancers can decrease dramatically from 25\%-44\% at localized stage to 3\% at the distant stage~\cite{liu2016prognosis,singal2022hcc,dbouk2022multicenter,siegel2025cancer,siegel2024cancer,siegel2026cancer,llovet2021hepatocellular}. As shown in \Cref{fig:overview}b, these cancers are upper gastrointestinal malignancies with shared pathological characteristics and therefore may present similar symptoms and are suitable for screening in a unified manner. To the best of our knowledge, HOPE ({H}epat{O}-{P}ancreato-biliary cancer screening with self-{E}volving trustworthy AI) is the first effective population-scale screening approach for HPB cancers.

{\bf To reduce false positives in population-scale screening, we introduce a three-stage risk-stratification funnel that progressively concentrates screening efforts on higher-risk candidates via more precise assessments, enabling a radical suppression of false positives while preserving sensitivity}. In {Stage 1}, HOPE identifies an {\bf at-risk stratum} with abnormal clinical indicators, which applies to hospital scenarios as well as community scenarios via questionnaires (Supplementary Information). In {Stage 2}, HOPE uses LLMs to analyze routine EHRs and predicts the {\bf high-risk stratum} among at-risk candidates, enabling physicians to determine whether patients require recall for non-contrast computed tomography (CT) scans. In {Stage 3}, HOPE detects the {\bf suspected stratum} by leveraging paired EHRs and non-contrast CTs using a multi-modal AI system, assisting physicians in prioritizing confirmatory HPB cancer diagnostics.

{\bf To enhance the model generalizability and interpretability in EHR data analysis, we propose a self-evolving algorithm~\cite{silver2017mastering,zelikman2024star,guo2025deepseek} to incentivize the reasoning capability of LLMs~\cite{wei2022chain}}. Without human annotations, this algorithm iteratively improves the reasoning capability and final accuracy by training on progressively higher-quality data, which are generated by the model in the previous iteration (\Cref{fig:overview}c). These data are rigorously filtered by a strict LLM verifier that assesses {answer correctness}, {reasoning-answer consistency}, and {truthfulness of reasoning processes}. We initialized from a general-purpose reasoning LLM and applied this self-evolving framework to enhance its reasoning capability for HPB cancer screening.

We established a cohort consisting of 10,269,153 patients, one of the largest cohorts for multi-cancer screening, for HOPE development and internal evaluation. Then, we conducted extensive evaluations to test its capability (\Cref{fig:overview}d). Our experiments highlighted several key advantages of HOPE. {\bf First}, HOPE substantially reduces false positives. The three-stage screening funnel achieved a significantly higher extrapolated PPV of 80.7\% than SOTA screening approaches (10\%-49\%)~\cite{liu2020sensitive,cao2023large,nadauld2021pathfinder,schrag2023pathfinder} while maintaining a high sensitivity. {\bf Second}, the reasoning capability significantly improves the model's generalizability. In external validation cohorts, the reasoning model achieved areas under the receiver operating characteristic curve (AUCs) of 0.937 and 0.928, significantly outperforming the non-reasoning baselines with AUCs of 0.745 and 0.777. {\bf Third}, the reasoning processes foster physician trust. Incorporating reasoning capabilities into the model increased the mean physician acceptance rate from 16.6\% to 64.1\%. {\bf Finally}, HOPE is cost-effective compared with SOTA single-test approaches. HOPE achieves greater quality-adjusted life years at lower cost in our health economics microsimulation experiment. In sum, HOPE could serve as the first effective population-scale screening approach for HPB cancers, and its generalizable design holds promise for extension to a broader spectrum of cancers (Supplementary Information).

\section{Results}
\label{sec:result}

\subsection{Overview of HOPE}
\label{subsec:hope}

HOPE is designed as a three-stage risk-stratification screening funnel for highly accurate population-scale HPB cancer screening, as shown in \Cref{fig:overview}a. We developed HOPE using a large-scale internal cohort of 10,269,153 candidates from the Affiliated Hospital of Qingdao University (AHQU), a major medical center in Shandong, China. The data consist of outpatient, inpatient and emergency patients. We applied strict inclusion and exclusion criteria to retain eligible patients. To ensure rigorous labels, we only included patients with confirmed HPB cancers or two-year negative follow-up periods. To ensure sufficient longitudinal clinical history, we included only individuals with at least three years of EHR data and a minimum of five clinical visits. Patients without adequate records are typically transferred from other hospitals and therefore should be screened at their primary hospital. For input prompt construction, to focus on broadly available routine clinical information, we excluded tumor biomarkers from the model inputs, such as CA19-9~\cite{ballehaninna2011serum} and AFP~\cite{tzartzeva2018surveillance}. Next, we describe the HOPE screening funnel in detail.

\begin{figure}[t]
    \centering
    \includegraphics[width=\linewidth]{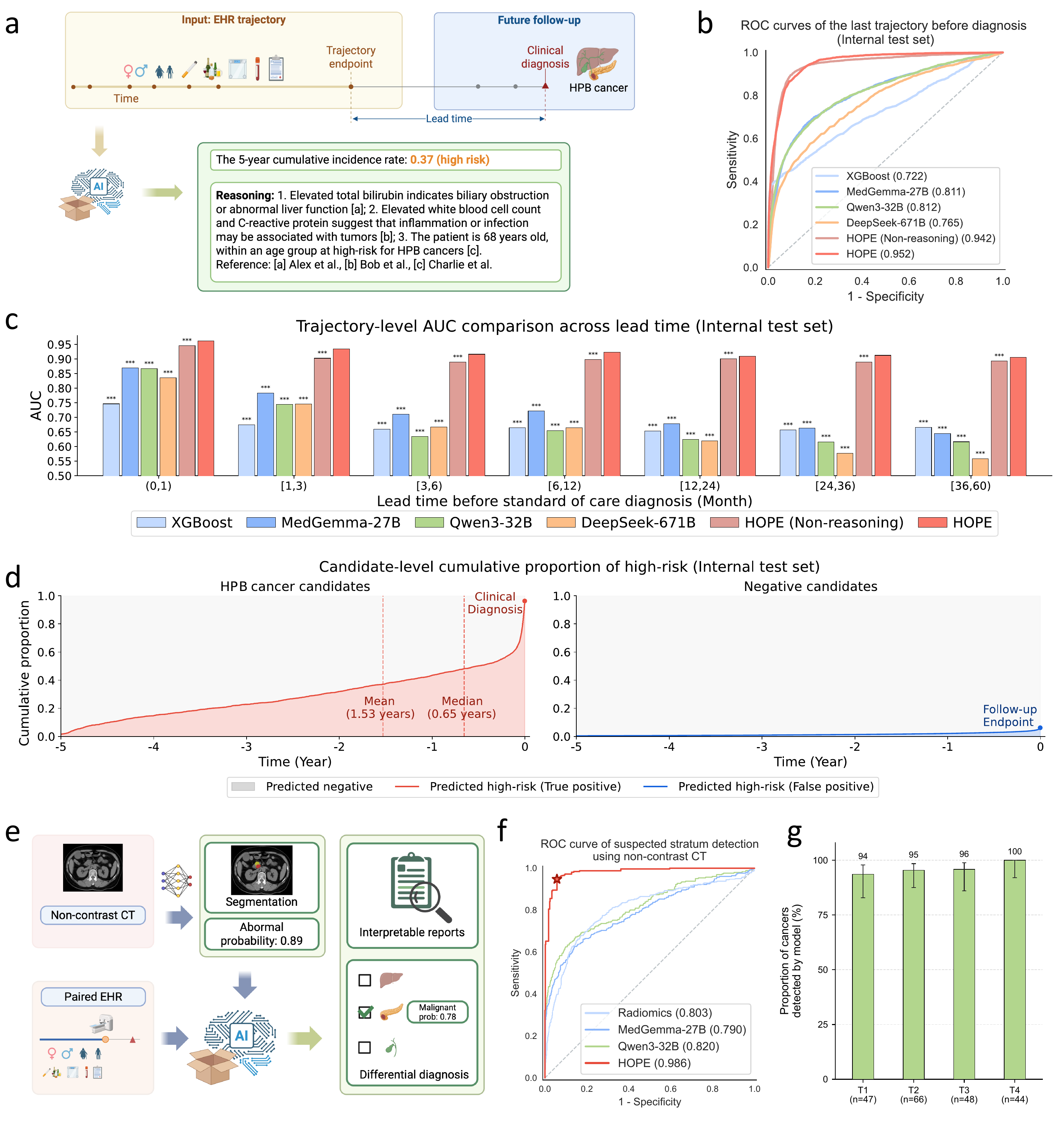}
    \caption{\small \small {\bf The results of internal evaluation.} {\bf a}, The schematic diagram of high-risk stratum prediction with the trustworthy AI. The lead time is defined as illustrated in this figure.
    {\bf b}, The receiver operating characteristic (ROC) curves of models for trajectory-level HPB cancer risk prediction using EHR data. 
    {\bf c}, Model performance across the lead-time intervals.
    {\bf d}, Proportion of high-risk candidates over time.
    {\bf e}, The schematic diagram of suspected stratum detection by the multi-modal AI system.
    {\bf f}, ROC curves for suspected stratum detection.
    {\bf g}, Proportion of HPB cancers detected by the multi-modal AI system in terms of T stage in the internal test cohort (n=205).*** P-value $<$ 0.001.}
    \label{fig:internal}
\end{figure}

In Stage 1, HOPE identifies an {\bf at-risk stratum} from the whole population using routine clinical indicators, including lifestyle~\cite{klein2021pancreatic,weissman2020diverse}, routine blood tests~\cite{takaaki2021clinical,xia2016advances,zhang2020albumin,wolpin2013hyperglycemia,hansen2021low,yuan2025liver}, and HPB cancer-related diseases~\cite{al2013improvement,yuan2025liver,molina2021deciphering,roy2021diabetes,hansen2021low,marengo2016liver,mcglynn2021epidemiology} (Methods~\Cref{subsubsec:stratum1}). Individuals are identified as at-risk if any single indicator is abnormal. This simple design does not rely on AI models, facilitating seamless integration into questionnaire-based workflows in community settings as well as in hospital information systems. This step preserves the majority of HPB cancer cases while substantially reducing the number of individuals requiring additional data acquisition and AI model evaluation, thereby lowering the overall cost of HOPE. We evaluated the performance of this stage using sensitivity, specificity, PPV, and negative predictive value (NPV). 
%TODO
% In the Supplementary Information, we also evaluated a simplified version of HOPE that omitted Stage 1 and applied Stages 2 and 3 directly to the entire population.

In Stage 2, HOPE predicts the {\bf high-risk stratum} using EHRs of at-risk candidates. The EHR data include clinical situations and their dynamic changes, encompassing demographics, lifestyle, body mass index (BMI), blood tests and diagnoses, etc. We construct an EHR trajectory with temporal information in free-form text to serve as the LLM prompt. HOPE attempts to predict the 5-year cumulative incidence rate starting from the trajectory endpoint as well as provide the diagnostic reasoning processes. To evaluate the model performance, we denote the ``lead time" as the time interval between trajectory endpoints and clinical endpoints (i.e., HPB cancer diagnoses for positive candidates or follow-up endpoint for negative candidates). The performance of models was evaluated at both the trajectory-level and the candidate-level. For trajectory-level evaluation,  we calculated the AUC using the last trajectory endpoint before the clinical endpoints, as well as reported AUCs by sampling positive trajectory endpoints at different lead-time intervals (i.e., 0-1, 1-3, 3-6, 6-12, 12-24, 24-36, and 36-60 months). For candidate-level evaluation, we calculate the high-risk proportion of positive candidates and negative candidates across lead time, respectively. We would proactively recommend non-contrast CT examinations for high-risk individuals without retrospective CT exams.

In Stage 3, HOPE detects the {\bf suspected stratum} by leveraging non-contrast CTs and paired EHRs of high-risk candidates. To analyze these data, we utilized a multi-modal AI system that consists of an nnUNet~\cite{cao2023large,isensee2021nnu} and a vision-language model (VLM)~\cite{bai2025qwen3}. This system detects organs and lesions, conducts differential diagnoses of cancer types, and provides interpretable reports. By providing detailed analyses, the system finally assists physicians in making final clinical decisions to confirm HPB cancers, such as surgery, biopsy, or follow-up. For evaluation, we utilized AUC, sensitivity, specificity, PPV, and NPV. Also, we calculated the proportion of HPB cancers detected by the multi-modal AI system across different American Joint Committee on Cancer (AJCC) T stages.

\begin{figure}[t]
    \centering
    \includegraphics[width=\linewidth]{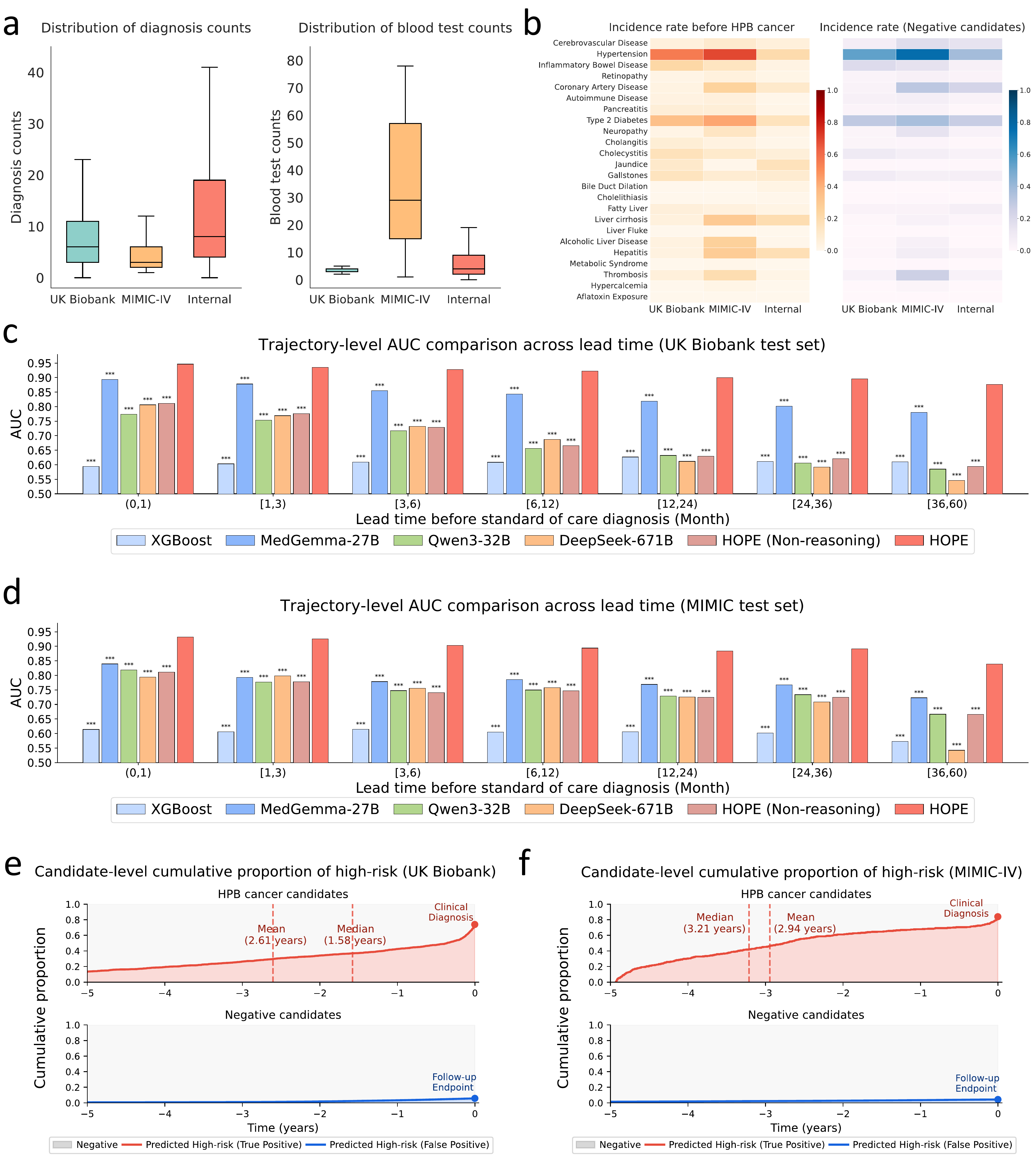}
    \caption{\small \small {\bf The results of external evaluation.} 
    {\bf a}, The distribution of diagnosis and blood test counts across internal and external datasets.
    {\bf b}, The comorbidity of HPB-cancer-related diseases across internal and external datasets.
    {\bf c-d}, Model performance across the lead-time intervals on UK Biobank and MIMIC-IV test sets.
    {\bf e-f}, Proportion of high-risk candidates over time on UK Biobank and MIMIC-IV test sets. *** P-value $<$ 0.001.
    }
    \label{fig:external}
\end{figure}

\subsection{Screening funnel reduces false positives}
\label{subsec:internal_eval}

We tested the effectiveness of each stratum on the internal cohort. After strict data inclusion and exclusion (Methods~\Cref{subsubsec:internal_data} and Supplementary Information), we retained 3,132,529 eligible candidates, consisting of 3,973 HPB cancer patients with an incidence rate of 0.13\%. Then, in Stage 1, we applied the at-risk stratification method using routine clinical indicators. In this step, we stratified 175,042 at-risk candidates (5.59\% of all eligible candidates); the incidence rate increased to 1.72\%, while retaining 75.76\% of HPB cancer cases. This simple step efficiently concentrates the screening scope, and it exhibits high interpretability as the utilized indicators are related to HPB cancers with clinical evidence (Methods~\Cref{subsubsec:stratum1}).

In Stage 2, HOPE predicts a high-risk stratum with self-evolving trustworthy AI, as illustrated in \Cref{fig:internal}a. We compare our trustworthy AI model with baselines, including XGBoost, MedGemma-27B, Qwen3-32B and DeepSeek-671B, using the area under the receiver operating characteristic curve (AUC). The XGBoost model is trained on internal data, and other baselines are open-source LLMs (Methods~\Cref{subsubsec:stratum2}).
As shown in \Cref{fig:internal}b, the AUC of HOPE is 0.952 (95\% CI 0.950-0.953), significantly surpassing the baselines (P-value $<$ 0.0001). In \Cref{fig:internal}c, we illustrate trajectory-level model performance across lead time. HOPE achieved an AUC of 0.962 (95\% CI 0.961-0.963) at a lead time within one month. As the lead time increases, the performance of baselines degrades substantially, whereas HOPE exhibits a modest decline. Notably, HOPE maintained an AUC of 0.905 (95\% CI 0.904-0.907) with the lead time between 36 and 60 months, while DeepSeek-671B only achieved 0.559 (95\% CI 0.557-0.561). In \Cref{fig:internal}d, we demonstrate the candidate-level high-risk prediction results over time. For individuals with HPB cancer, the proportion of high-risk candidates (i.e., sensitivity) consistently increases over time, while the false positive rate remains persistently low among negative individuals. Notably, the average lead time was 1.53 years, indicating that a substantial proportion of candidates could potentially be identified and treated at an earlier stage. For candidate-level high-risk prediction, the model achieved a sensitivity of 96.3\% and a specificity of 93.4\%. The PPV of HOPE is 20.3\%, and the negative predictive value (NPV) is 99.9\%. The results indicate that the cumulative HPB cancer incidence rate of the high-risk population is 20.3\%.

In Stage 3, we utilized a multi-modal AI system to detect suspected candidates using paired non-contrast CT and EHR data, as shown in \Cref{fig:internal}e. This system achieved an AUC of 0.986 (95\% CI 0.980-0.991) for HPB cancer detection from high-risk candidates, with a sensitivity of 94.8\% and a specificity of 94.2\%, outperforming SOTA baselines such as Radiomics, MedGemma-27B and Qwen3-32B (\Cref{fig:internal}f). For differential diagnosis, the AUC of liver cancer is 0.989 (95\% CI 0.980-0.992), pancreatic cancer is 0.987 (95\% CI 0.984-0.993), and biliary cancer is 0.876 (95\% CI 0.861-0.890). Importantly, our multi-modal AI maintained high detection proportions across different T stages. The proportions of T1, T2, T3, and T4 tumors are 93.6\% (95\% CI 82.8\%-97.8\%), 95.5\% (95\% CI 87.5\%-98.4\%), 95.8\% (95\% CI 86.0\%-98.8\%), and 100.0\% (95\% CI 92.0\%-100.0\%), demonstrating its ability to detect HPB cancers across a broad spectrum of disease progression (Fig. 2g). Notably, the overall screening funnel achieved an extrapolated PPV of 80.7\% while maintaining a high extrapolated sensitivity of 69.2\%, outperforming the SOTA single-test approaches.

\begin{figure}[t]
    \centering
    \includegraphics[width=\linewidth]{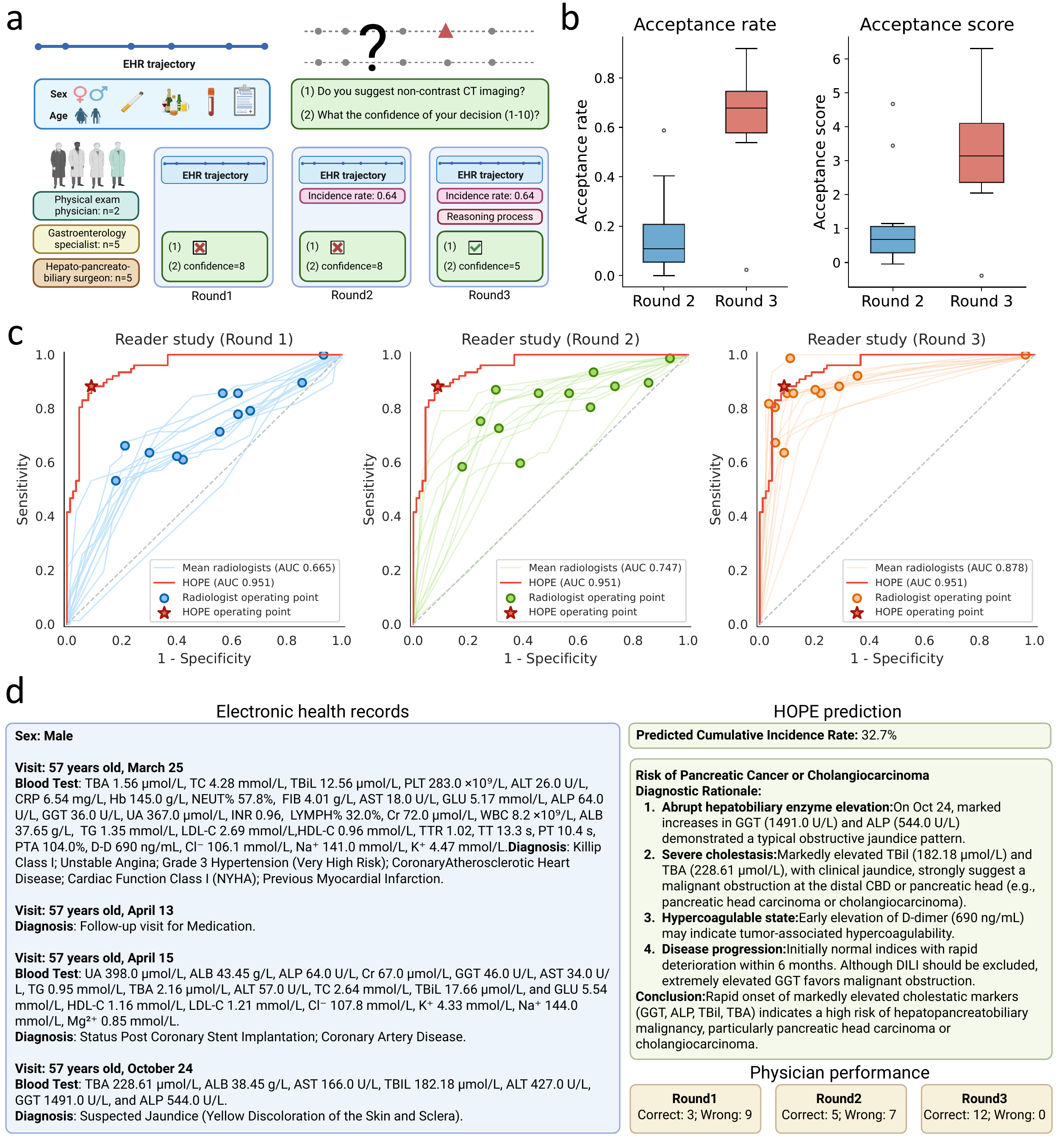}
    \caption{\small \small {\bf The results of reader study.} {\bf a}, Overview of the reader study. We compared HOPE with physicians and assessed whether the reasoning processes can foster physician trust. {\bf b}, The acceptance rate and acceptance score. {\bf c}, ROC curves and operating points of HOPE and readers. {\bf d}, Showcase of HOPE fostering physician trust. For concise presentation, references in the reasoning process were omitted.}
    \label{fig:reader}
\end{figure}

\subsection{Reasoning capability improves generalizability}
\label{subsec:external_eval}

We evaluated the generalization ability of HOPE Stage 2 on two external datasets, UK Biobank~\cite{sudlow2015uk} and MIMIC-IV~\cite{johnson2023mimic}. The UK Biobank is a prospective cohort of United Kingdom candidates with longitudinal follow-up clinical data, and MIMIC-IV is an inpatient cohort collected in a hospital in the United States. We observed distribution shifts between the internal and external datasets, primarily stemming from differences in clinical settings and data collection methodologies. The distributions of diagnosis counts and blood test counts were different (\Cref{fig:external}a), and the comorbidity patterns of HPB-cancer-related diseases also differed across these datasets (\Cref{fig:external}b). We applied the inclusion and exclusion criteria to these cohorts. This step retained 389,103 eligible candidates in the UK Biobank (n=501,946) and 63,316 eligible candidates in MIMIC-IV (n=364,627). We identified 149,363 at-risk candidates in the UK Biobank, and 39,016 at-risk candidates in MIMIC-IV. More details of at-risk stratification are listed in Supplementary Information.

In Stage 2, our trustworthy AI model achieved a trajectory-level AUC of 0.937 (95\% CI 0.935–0.938) on the UK Biobank and 0.928 (95\% CI 0.925–0.931) on MIMIC-IV using the last trajectory before diagnosis, significantly outperforming all baseline models (P-value $<$ 0.001; \Cref{fig:external}c,d). Importantly, our model demonstrated sustained predictive performance over extended horizons, maintaining AUCs of 0.877 (95\% CI 0.874–0.879) on UK Biobank and 0.839 (95\% CI 0.835–0.843) on MIMIC-IV for lead times of 36–60 months. Although the non-reasoning LLM achieved comparable performance to the reasoning LLM on the internal test set, its performance declined substantially on external cohorts, with AUCs of 0.745 (95\% CI 0.742–0.748) on UK Biobank and 0.777 (95\% CI 0.772–0.782) on MIMIC-IV. The limited generalizability of non-reasoning LLMs has also been observed in previous studies~\cite{placido2023deep}. In contrast, the reasoning LLM maintained robust performance across all external cohorts, suggesting that explicit reasoning may promote the learning of more transferable clinical representations and enhance generalizability across heterogeneous healthcare settings.

For candidate-level evaluation of HOPE Stage 2, we assessed the longitudinal risk prediction performance of HOPE across external test cohorts (\Cref{fig:external}e,f). In the UK Biobank, HOPE achieved a sensitivity of 74.0\%, specificity of 94.3\%, PPV of 12.18\%, and NPV of 99.7\%. In MIMIC-IV, HOPE demonstrated comparable performance, achieving a sensitivity of 83.9\%, specificity of 95.7\%, PPV of 20.1\%, and NPV of 99.8\%. Notably, HOPE identified individuals at high risk of HPB cancer with an average lead time of 2.61 years in the UK Biobank and 2.94 years in MIMIC-IV, demonstrating its potential for early cancer detection and risk stratification across independent healthcare settings.

To evaluate the clinical utility of HOPE, we conducted a real-world study aimed at identifying HPB cancer cases missed by the standard of care (SoC). The study was performed using a continuous patient cohort from Yantai Yuhuangding Hospital (YHD), a major medical center in Shandong, China. We included 139,373 patients who visited YHD between 1 May and 31 May 2025, with longitudinal EHRs extracted from the hospital information system (HIS). By 1 May 2026, 577 patients had confirmed HPB cancer diagnoses, including 148 patients diagnosed after 31 May 2025. Beyond these confirmed cases, HOPE identified 43 high-risk candidates and 22 suspected patients who had not been identified through the SoC pathway. We contacted these 22 suspected patients for follow-up verification and invited them to undergo additional diagnostic evaluation. One patient returned to the hospital for additional assessment, which confirmed the presence of liver cancer. For the other suspected patients, five could not be reached, nine declined further evaluation, three had died with unconfirmed HPB tumors and four had died with HPB cancer.

\subsection{Reasoning processes foster physician trust}
\label{subsec:reader_study}

We conducted a reader study to compare HOPE with physicians and to assess whether HOPE's reasoning processes effectively foster physician trust. Given the EHR trajectory, readers evaluated the necessity of a non-contrast CT examination and their level of confidence in that assessment. Twelve board-certified physicians participated in the reader study, with clinical experience ranging from 1 to 18 years, including physical examination physicians (n=2), gastroenterology specialists (n=5), and HPB surgeons (n=5). The reader study included 167 EHR trajectories, comprising 77 HPB cancer cases and 90 negative cases. As shown in \Cref{fig:reader}a, the reader study consists of three assessment rounds. In Round 1, readers independently assessed HPB cancer risk using EHR data alone and determined whether a non-contrast CT examination was warranted, establishing a baseline measure of clinical performance. In Round 2, readers were additionally provided with HOPE-predicted HPB cancer risk probabilities and revised their assessments accordingly. In Round 3, readers received both the predicted probabilities and the model's underlying reasoning process, upon which their final assessments were based. Across all three rounds, readers rated their confidence in each prediction on a scale of 1 to 10.

We demonstrate that the reasoning process of HOPE meaningfully enhances physician trust in model predictions. The physician acceptance rate is defined as the proportion of cases in which physicians revised their initial decisions to align with the model when disagreement arose; the acceptance score is defined as the average change in confidence toward the model's prediction. As shown in \Cref{fig:reader}b, the mean acceptance rate in Round 2 remains low (16.6\%), reflecting limited trust in a black-box model lacking interpretable outputs. Providing the reasoning process in Round 3 increased the acceptance rate to 64.1\%. Consistently, the median acceptance score improved from 0.68 to 3.14. Inter-rater agreement, measured by Fleiss' kappa, was 0.228 in Round 1, 0.237 in Round 2, and 0.477 in Round 3, reflecting a substantial improvement in decision consistency.

Reader performance varied in accordance with the degree of physician trust, as illustrated in \Cref{fig:reader}c, where ROC curves of readers were calculated from confidence scores of responses. In Round 1, HOPE substantially outperformed independent physician predictions, achieving an AUC of 0.951 compared with an average physician AUC of 0.665. In Round 2, physician performance improved modestly to an average AUC of 0.747, yet remained significantly below the model. In Round 3, provision of the model's reasoning process led to a marked improvement, with the average physician AUC increasing to 0.878. Moreover, physicians' operating points shifted significantly toward the upper-left region, indicating substantially improved decision consistency with the model.

A representative case is shown in \Cref{fig:reader}d, demonstrating the capacity of the model's reasoning process to correct erroneous physician decisions. For this case, HOPE predicted a 5-year HPB cancer risk of 32.7\%, as shown in the figure. In Round 1, only three of twelve physicians made the correct judgment when working independently. The provision of predicted probabilities alone in Round 2 led to modest improvement, with five of twelve readers reaching the correct decision, reflecting persistently limited trust in the model. In Round 3, provision of the reasoning process resulted in complete consensus, with all twelve physicians aligning with the model's recommendation.

\begin{figure}[t]
    \centering
    \includegraphics[width=\linewidth]{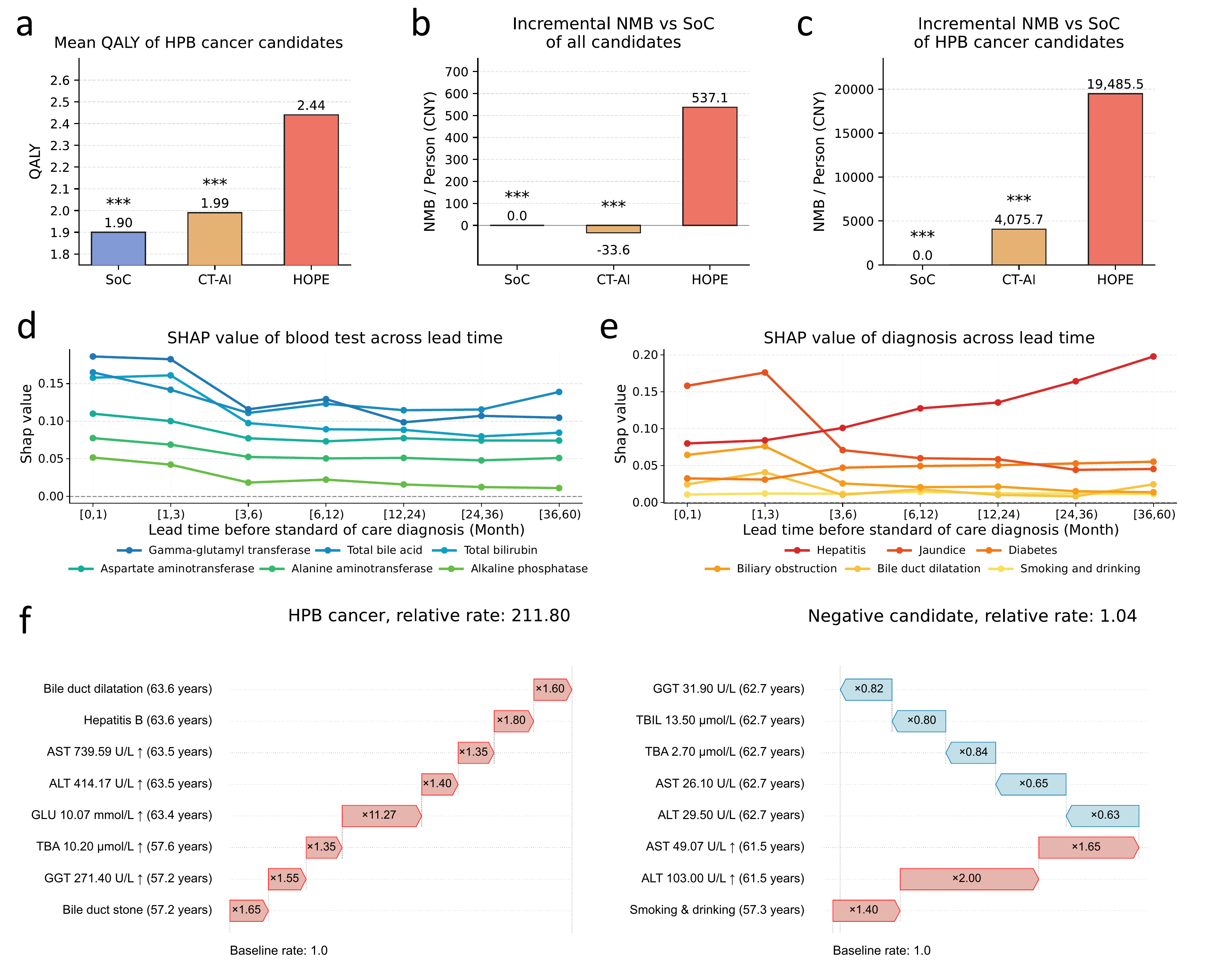}
    \caption{\small \small {\bf HOPE is cost-effective and interpretable.} {\bf a}, Mean post-cancer-onset QALYs among strategy-specific confirmed HPB cancer cases under SoC, CT-AI, and HOPE. {\bf b}, Incremental NMB of all candidates in comparison to SoC. {\bf c}, Incremental NMB of HPB cancer candidates in comparison to SoC. {\bf d}, The SHAP values of blood tests across lead time. {\bf e}, The SHAP values of diagnosed diseases across lead time. {\bf f}, SHAP-explained risk contributions for EHR trajectories of an HPB cancer candidate and a negative candidate.}
    \label{fig:cost_eff}
\end{figure}

\subsection{HOPE is health-economically cost-effective}
\label{subsec:cost_efficiency}

To evaluate the economic value of HOPE for HPB cancer detection, we developed an individual-level microsimulation comparing three screening approaches: standard of care (SoC), AI-based opportunistic screening using non-contrast CT (CT-AI), and HOPE. The details of baselines are illustrated in Methods~\Cref{subsubsec:cost_effectiveness_methods}. We simulated a population of 500,000 candidates with age-related mortality and cancer incidence rates~\cite{sung2021global,wang2020global}. For simulated patients with HPB cancer, we modeled disease progression and treatment outcomes using stage-specific parameters informed by published survival and treatment evidence~\cite{llovet2022immunotherapies,park2021pancreatic,valle2021biliary}. We included the basic cost of SoC, as well as the costs of AI deployment, confirmatory diagnoses and treatment. We also included the additional costs of non-contrast CTs in HOPE. For evaluation, we use the quality-adjusted life-years (QALYs) to quantify the utility of cancer screening approaches and net monetary benefit (NMB) to quantify the socioeconomic benefit. All monetary outcomes below are reported in Chinese Yuan (CNY).

We demonstrate that HOPE significantly improved the mean QALY and incremental NMB compared with the SoC and CT-AI baselines. As shown in \Cref{fig:cost_eff}a, the mean QALY of HPB cancer candidates using HOPE is 2.44 (95\% CI 2.36-2.53), significantly surpassing the SoC (1.90, 95\% CI 1.83-1.97) and CT-AI (1.99, 95\% CI 1.92-2.06). Note that these QALYs describe strategy-specific confirmed cases rather than a common cohort; results for the common incident-cancer cohort are provided in Supplementary Information. As shown in \Cref{fig:cost_eff}b, we calculate the incremental NMB with the baseline of SoC, using a willingness-to-pay threshold of 287,247 per QALY (three times China's per capita GDP). For all candidates, HOPE achieved 537.1 (95\% CI 480.7-594.7) incremental NMB relative to SoC. Note that the CT-AI only achieved $-33.6$ (95\% CI $-54.8$ to $-11.5$), as the monetary benefits may be insufficient to cover the costs of AI deployment and false positives. As shown in \Cref{fig:cost_eff}c, for HPB cancer candidates, HOPE achieved 19,485.5 (95\% CI 16,961.9-22,074.0) incremental NMB relative to SoC, and significantly outperformed the CT-AI approach.

\subsection{Explaining AI predictions}
\label{subsec:explain}

To interpret the predictions of the trustworthy AI, we calculated the SHAP values of blood tests and diagnoses across the lead time. We first examined how the contributions of blood test indicators varied before the HPB cancer diagnosis. In \Cref{fig:cost_eff}d, we show how the SHAP values of the most influential blood-test indicators vary across the lead time. We indicate that hepatobiliary biomarkers consistently exhibited the largest SHAP values, including gamma-glutamyl transferase (GGT), total bile acids (TBA), and total bilirubin (TBIL), whereas aspartate aminotransferase (AST) and alanine aminotransferase (ALT) showed moderate but persistent contributions. As shown in \Cref{fig:cost_eff}e, hepatitis showed progressively greater contributions with increasing lead time, suggesting a long-term association with cancer risk, whereas jaundice contributed most strongly shortly before HPB cancer diagnosis, indicating a short-term signal.

Further, we respectively showcase the SHAP-explained risk contributions for EHR trajectories of an HPB cancer candidate and a negative candidate in \Cref{fig:cost_eff}f. For the HPB cancer candidate, the explanation framework identified elevated glucose as the largest positive contribution, while bile duct stones, hepatitis B, bile duct dilatation, and elevated levels of GGT, TBA, ALT and AST also made substantial contributions to the predicted risk.
For the negative candidate, positive contributions associated with the lifestyle of smoking and drinking, as well as elevated ALT and AST in an earlier period, are offset by subsequent reductions in ALT and AST and normal GGT, TBIL and TBA. These cases illustrate the progressive accumulation of cancer risk in an HPB cancer candidate, as well as how the model lowers the estimated risk as clinical indicators return to normal in a negative candidate, thereby avoiding a false-positive prediction.

\section{Discussion}
\label{sec:discussion}

Cancer screening is challenging to implement at a population scale, as current single-test approaches~\cite{nabavizadeh2025lba64,nadauld2021pathfinder} can lead to a large number of false positives due to the low prevalence. At present, recommended screening is only established for a few cancers, such as colorectal, female breast, cervical, and smoking-associated lung cancers. For other cancers, screening remains infeasible or restricted to a limited scale, such as opportunistic scenarios, symptomatic individuals, or patients with family history. As a result, the current implementation of screening covers only a minority of cancer patients. For example, more than 80\% of pancreatic cancers are detected at a distant stage. This challenge largely constrains the potential utility of screening.

To facilitate population-scale screening, we propose a risk stratification funnel to achieve high accuracy, clinical feasibility and cost efficiency. The first two Stages of the funnel elegantly leverage data from routine clinical care, thereby circumventing the need for dedicated, high-cost examinations. Only high-risk individuals are selected to undergo further examinations, such as non-contrast CT scans. This progressive concentration of screening efforts enables a radical suppression of false positives while preserving sensitivity. On our internal cohort, HOPE achieved an extrapolated PPV of 80.7\% and an extrapolated sensitivity of 69.2\%, substantially outperforming current single-test screening approaches.

Beyond the high performance of AI models, the clinical utility of AI is fundamentally predicated on physician trust. To foster physician trust, HOPE employs a self-evolving algorithm to enhance LLMs' diagnostic reasoning capability. Our reader study quantifies this impact: the diagnostic reasoning catalyzed a nearly four-fold increase in acceptance from 16.6\% to 64.1\%, and a significant elevation in inter-rater concordance. Crucially, the collaborative synergy between AI reasoning and human expertise boosted the average physician AUC from 0.665 to 0.878, nearing the model's standalone performance. By investigating the successful cases, we find that HOPE is able to uncover HPB cancer-related indicators that were overlooked by readers.

\section{Methods}
\label{sec:methods}

\subsection{Ethics approval}

Our study was approved by the institutional review board of the Affiliated Hospital of Qingdao University (ID: QYFYEC2025-33) and Yantai Yuhuangding Hospital (ID: K2026-336). The study was not interventional and was performed under guidelines approved by the institutional review board. Informed consent was waived since the study presents no more than minimal risk. All data in this study were de-identified prior to model training, testing and reader studies.

\subsection{Dataset description}
\label{subsec:dataset_description}

To develop and evaluate the screening approach, we utilized four cohorts: an internal cohort from the Affiliated Hospital of Qingdao University (AHQU) for model development and internal evaluation; two external cohorts, UK Biobank and MIMIC-IV, to assess the generalizability of the models; and a real-world cohort from the Yantai Yuhuangding Hospital (YHD) to evaluate the clinical utility of HOPE. HPB cancer is defined as the primary carcinoma of the liver, pancreas, and biliary system. We carefully excluded patients with secondary HPB malignancies or pathologically unconfirmed HPB tumors, but retained patients with confirmed benign diseases. We used the International Classification of Diseases, 10th Revision (ICD-10) coding system to standardize all diagnoses. To facilitate the multi-modal AI system of HOPE, we retrospectively collected the non-contrast CT scans of high-risk candidates in the AHQU and YHD cohorts. Furthermore, we collected the available HPB cancer staging information from pathology reports according to the eighth edition of the American Joint Committee on Cancer (AJCC) cancer staging system.

\subsubsection{Internal cohort}
\label{subsubsec:internal_data}

The internal cohort comprised 10,269,153 patients who visited AHQU between January 2012 and May 2025. Following strict inclusion and exclusion criteria, we retained 3,132,529 eligible candidates, of whom 3,973 were confirmed to have HPB cancer. Then, we stratified 175,042 at-risk candidates based on abnormal routine clinical indicators. For the development and evaluation of AI models, we randomly split the at-risk candidates into a training-validation set (70\%) and an independent test set (30\%). Then, the AI model predicted 14,253 high-risk candidates via 5-fold cross-validation and independent testing. We then retrospectively retrieved non-contrast CT scans for 6,728 of these high-risk candidates; the remaining individuals did not undergo CT imaging at AHQU. Within this subset, 1,376 patients had confirmed HPB cancer, 205 of whom had available AJCC T-stage information.

\subsubsection{External cohorts}
\label{subsubsec:external_data}

We used UK Biobank and MIMIC-IV as external cohorts to evaluate the generalizability of HOPE. UK Biobank is a prospective cohort comprising 501,946 candidates recruited in the United Kingdom. After applying the inclusion and exclusion criteria, we retained 389,103 candidates, including 2,436 candidates with HPB cancer. MIMIC-IV is an inpatient cohort comprising 364,627 patients who visited Beth Israel Deaconess Medical Center between 2008 and 2022. Following the inclusion and exclusion criteria, we retained 63,316 patients, including 513 patients with HPB cancer. The distributions in UK Biobank and MIMIC-IV differ significantly from those in the hospital-collected AHQU data, as shown in \Cref{fig:external}a, \Cref{fig:external}b and Supplementary Information. Notably, because UK Biobank candidates underwent a maximum of four blood tests, many underlying abnormalities in routine clinical indicators likely went undetected. In contrast, as the MIMIC-IV cohort consists of inpatients, clinical abnormalities are present in nearly all candidates.

\subsubsection{Real-world cohort}
\label{subsubsec:real_world}

The real-world cohort comprised 139,373 consecutive patients who visited YHD between 1 and 31 May 2025. By 1 May 2026, 577 patients had a confirmed diagnosis of HPB cancer, of whom 148 were diagnosed with HPB cancer after 31 May 2025. We used this cohort to assess the capability of HOPE to identify HPB cases missed by SoC in a real-world clinical setting.

\subsection{HOPE approach}
\label{subsec:HOPE}

\subsubsection{Stage 1: At-risk stratum identification}
\label{subsubsec:stratum1}

In Stage 1, we used 18 routine clinical indicators, including 13 blood-test indicators (Glycated Hemoglobin, Total bilirubin, Total Bile Acid, Aspartate Aminotransferase, Alanine Aminotransferase, Gamma-Glutamyl Transferase, Alkaline Phosphatase, Amylase, Prothrombin Time, D-Dimer, Creatinine, Fasting Blood Glucose, Sodium), 4 HPB-related diseases (Diabetes, Liver cirrhosis, Pancreatitis and Jaundice), and 1 composite lifestyle indicator (smoking and alcohol consumption). We identified at-risk candidates when they were 45 years of age or older and triggered any abnormal blood-test indicator, any of the HPB-related diseases, or the lifestyle habit. The missing blood tests, diagnoses or lifestyle information were treated as normal.

The index date for the at-risk status, denoted as abnormal date, was defined as the earliest recorded instance of an abnormal indicator. If this date preceded the age of 45, it was set to the date the candidate reached 45 years of age. Individuals whose abnormal date postdated their HPB cancer diagnosis were treated as false negatives missed by the HOPE screening funnel. Importantly, for identified at-risk candidates, the abnormal date significantly preceded the HPB cancer diagnosis. The average lead time was 2.55 years in the internal cohort, 6.3 years in the UK Biobank, and 6.6 years in the MIMIC-IV dataset. More details are provided in the Supplementary Information.

\subsubsection{Stage 2: High-risk stratum prediction}
\label{subsubsec:stratum2}

In Stage 2, we used a self-evolving trustworthy AI model $\mathrm{M}_\mathrm{\theta}$ to predict the probability of HPB cancer in 5 years, $\mathrm{P}_{\text{risk}}$, as well as the diagnostic reasoning processes $\mathrm{R}$ from the input query of EHR trajectories $\mathrm{Q}$. We constructed the input query of EHR trajectories as free-form text with a predefined format. For candidates with HPB cancer, we sampled trajectory endpoints before the diagnostic visit (e.g., observation of an unconfirmed HPB mass). For negative candidates, we sampled trajectory endpoints after the abnormal date. To confirm the negative labels, we required at least 2 years of follow-up without HPB cancer after the endpoints. Without human-defined reasoning processes, we designed a self-evolving algorithm to incentivize the diagnostic reasoning capability. 

This algorithm attempts to iteratively train the model with higher-quality data to generate better reasoning processes, as shown in \Cref{fig:overview}c. We started from an open-source general-purpose LLM $\mathrm{M}_\mathrm{\theta}^{0}$ to generate the reasoning process $\mathrm{R}^0$ and the answer $\mathrm{A}^0$ for each query $\mathrm{Q}$. Then, we applied the LLM verifier to assess the correctness of the generated reasoning processes and answers. After data verification, we trained the $\mathrm{M}_\mathrm{\theta}^{0}$ model using the verified dataset. For the i-th iteration, we conducted data generation using $\mathrm{M}_\mathrm{\theta}^{i}$, performed data verification, and trained the model from $\mathrm{M}_\mathrm{\theta}^{0}$, where $\text{i}=1,2,3$. To prevent performance saturation during the iterative evolution, we provide the correct answer as a hint to the model $\mathrm{M}_\mathrm{\theta}^{i}$ during the generation of $\mathrm{R}^{\text{i}}$ and $\mathrm{A}^{\text{i}}$ from $\mathrm{Q}$ when ${\text{i}}=2,3$. This design was inspired by the STaR algorithm~\cite{zelikman2024star} and will not cause information leakage, as we do not provide answers during the training period. After three iterations, we get the final model $\mathrm{M}_\mathrm{\theta}=\mathrm{M}_\mathrm{\theta}^{3}$. During the inference, for each input query $\mathrm{Q}$ of EHR trajectory, we get the risk score,
$$
\mathrm{S}_{\text{risk}}=\frac{\exp\left(\mathrm{logprob}_{\mathrm{M}_{\mathrm{\theta}}}(\mathrm{A}_{\text{pos}}|\mathrm{R},\mathrm{Q})\right)}{\sum_{\mathrm{A}\in \mathcal{A}}\exp\left(\mathrm{logprob}_{\mathrm{M}_{\mathrm{\theta}}}(\mathrm{A}|\mathrm{R},\mathrm{Q})\right)},
$$
where $\mathcal{A}=\{\mathrm{A}_{\text{pos}}, \mathrm{A}_{\text{neg}}\}$ and $\mathrm{logprob}_{\mathrm{M}_{\mathrm{\theta}}}(\cdot)$ is the average log probability of answer tokens. Finally, we applied Isotonic Regression to the predicted scores $\mathrm{S}_{\text{risk}}$ to get the predicted probability of HPB cancer $\mathrm{P}_{\text{risk}}$.
Detailed training procedures are provided in the Supplementary Information.

\subsubsection{Stage 3: Suspected stratum detection}
\label{subsubsec:stratum3}

In Stage 3, we used a multi-modal AI system to detect the suspected candidates from the paired EHR and non-contrast CT data, as shown in \Cref{fig:internal}e. The multi-modal AI system consisted of an nnUNet and a vision-language model (VLM). Using the non-contrast CTs, we trained an nnUNet to segment the HPB organ and abnormal lesion masks, as well as to predict the abnormality probability, inspired by the PANDA algorithm~\cite{cao2023large}. Then, we constructed the paired EHR trajectory endpoint to be the same as the CT date. We input (1) the paired EHR and non-contrast CT, and (2) the nnUNet predictions into the VLM and trained it to output the interpretable report as well as the differential diagnosis of cancer types (i.e., liver cancer, pancreatic cancer and biliary cancer). The interpretable reports integrate analysis and reasoning based on both CT and EHR data. The annotations were constructed based on the correct elements of CT reports and the EHR reasoning processes. Finally, we provided the predicted results of organ and lesion masks, interpretable reports, and differential diagnosis to assist physicians in making final clinical decisions.

\subsection{Internal and external evaluation}
\label{subsec:internal_methods}

% done: YHJ
We extensively evaluated the HOPE approach using the internal cohort and two external cohorts (UK Biobank and MIMIC-IV). For Stage 1, we evaluated the utility of at-risk stratification across different screening scenarios. We concluded that its necessity depends on the clinical setting: this stage is particularly important for population-scale screening, where it can substantially reduce the number of individuals requiring downstream assessment. However, Stage 1 is useless in the inpatient setting, as most candidates were stratified as at risk, substantially limiting its potential for cost reduction and diminishing its practical necessity.

% To further demonstrate its utility in a population-scale setting, we show an additional evaluation using a cohort (n=1,259,933) from Yinzhou District, Ningbo, in the Supplementary Information. However, Stage 1 is useless in the inpatient setting, as most candidates were stratified as at risk, substantially limiting its potential for cost reduction and diminishing its practical necessity.

% done: YHJ
For Stage 2, we conducted trajectory-level and candidate-level evaluation to assess the model's capability of predicting high-risk candidates. For trajectory-level evaluation, we calculate the AUCs of sampled positive trajectories at different lead-time intervals. To assess the model performance of high-risk prediction before SoC diagnoses, we sampled each positive trajectory endpoint as the last visit before HPB cancer diagnosis. We define the diagnostic visit as the first detection of HPB tumors instead of pathological confirmation to prevent information leakage. To assess model performance of high-risk prediction across lead time, we sampled positive trajectory endpoints at different lead-time intervals. Under this setting, multiple trajectories may be generated for the same patient. For candidate-level evaluation, we calculate the high-risk proportion in positive and negative candidates across lead time. To account for the cumulative impact of false-positive findings, high-risk candidates were retained even when their predicted risk subsequently fell below the threshold. For positive candidates, we also calculated the mean and median values of lead time between HOPE's high-risk predictions and HPB cancer diagnoses.

% done: YHJ
For Stage 3, we evaluated the multi-modal AI system for suspected stratum detection using paired non-contrast CT scans and EHR trajectories. We calculated the AUC score for suspected candidate detection, and we also reported the AUC score for each cancer type. In the Supplementary Information, we showcase the input CT and EHR of an HPB cancer candidate as well as the output results of the AI system, including the segmentation masks, interpretable report, and differential diagnosis results. To assess the capability of the AI system for detecting early-stage cancers, we calculated the proportion of HPB cancers detected across AJCC T stages. As the UK Biobank and MIMIC-IV datasets lack CT scans of the candidates, the performance of this AI system was only evaluated on the internal test set.

\subsection{Reader study}
\label{subsec:reader_methods}

% done: YHJ
We conducted a reader study to (1) compare the performance of HOPE with that of physicians and (2) assess whether its reasoning process enhances physician trust. The reader study consisted of three assessment rounds: in Round 1, readers independently assessed the HPB cancer risks; in Round 2, readers were assisted by the AI-predicted risk probability; in Round 3, readers were further provided with the reasoning processes. For methodological rigor, the three assessment rounds were separated by a washout period of at least one month, and the case order was randomized for each reader in each round. Also, readers could not access or modify their answers in previous rounds. The reader study was conducted using a web-based interface specifically developed for this study, as detailed in the Supplementary Information.

% done: YHJ
For evaluation, we used the recommendations for proactive CT examinations to calculate evaluation metrics. We combined each CT recommendation with its confidence score to calculate the AUC for each reader in each round. To calculate acceptance rates, we focused on cases in which readers' recommendations differed from those of HOPE in Round 1 and calculated the proportion of cases in which readers changed their recommendations in Rounds 2 and 3. Further, we defined the acceptance score as the mean change in the readers' confidence towards the HOPE prediction. Note that the calculation of acceptance rates and scores considered physician trust only, irrespective of answer correctness. Inter-rater agreement was assessed separately for each round using Fleiss' kappa. We provide additional details on study implementation, reader prediction results and specialty-specific analyses in the Supplementary Information.

\subsection{Health economics microsimulation}
\label{subsubsec:cost_effectiveness_methods}

% done: YHJ
We conducted an individual-level microsimulation to compare the health-economic effects of SoC, CT-AI and HOPE. The simulated cohort included 500,000 candidates who entered the model at age 45 and were followed in annual cycles until death. We used age-specific distributions for HPB cancer incidence rates and background mortality. Also, we used the cancer stage-related life-utility multipliers, treatment, disease progression, and mortality. The approaches differed only in how suspected cancers were identified before confirmatory diagnosis. 

% done: YHJ
The SoC approach only included the diagnosis and treatment of routine medical conditions, and the cancer detection rates at each stage were derived from published literature. The CT-AI approach is an opportunistic screening approach, using non-contrast CTs (originally obtained for other reasons) to detect HPB cancers missed by physicians in the SoC. For the HOPE approach, only patients predicted to be positive at all three stages were considered positive.

% done: YHJ
We evaluated the approaches using QALYs and incremental NMB. We calculated QALYs using lifetime survival weighted by baseline utility and utility multipliers of HPB cancer, with a short-term utility decrement for false-positive diagnoses. QALYs were discounted at an annual rate of 5\% in the base-case analysis. We also calculated the incremental NMB of the HPB cancer candidates, regardless of whether their cancer was detected or confirmed under the screening approach.

\subsection{Explaining AI predictions}
\label{methods:interpretation}

% done: YHJ
To analyze feature contributions associated with predictions from long free-form EHR trajectories, we adopted a surrogate model-based explanation framework. The analysis included 235,031 eligible internal EHR trajectories. Specifically, we extracted handcrafted features of clinical indicators, trained a LightGBM~\cite{ke2017lightgbm} surrogate model to approximate this risk margin, and applied TreeSHAP~\cite{lundberg2020local} to estimate the attribution of each indicator. We define the LLM-derived risk margin as the difference between the log probabilities of the positive and negative categories. Note that SHAP values were derived from the surrogate model and therefore should not be interpreted as direct measures of causal effects, clinical associations, or calibrated estimates of clinical risk.

% done: YHJ
The explanation framework contains three main steps. First, we represented clinical indicators using 495 handcrafted features, encompassing presence, risk direction, frequency, severity, recency, and longitudinal change. Second, we trained a LightGBM model with a positive monotonic constraint to approximate the LLM-derived risk margins. We verify the fidelity of surrogate margins using $\text{R}^2$, Pearson correlation, mean absolute error and root mean squared error with the ground truth LLM-risk margin (Supplementary Information). Finally, we applied TreeSHAP to quantify feature contributions. We summed the signed SHAP values of features corresponding to the same clinical indicator and used the mean absolute grouped SHAP values to assess global and temporal importance across lead-time intervals. For the candidate case reports, we linked the SHAP values to the ages at which the corresponding clinical evidence was recorded. Details are illustrated in Supplementary Information.

\subsection{Statistical analysis}
\label{subsec:statistical_analysis}

All confidence intervals were reported at the 95\% level. To assess statistical significance, P-values between correlated receiver operating characteristic (ROC) curves were calculated using DeLong’s test. P-values from multiple comparisons were adjusted using the Benjamini–Hochberg (BH) method to control the false discovery rate (FDR) at 5\%. A two-sided P-value $<$ 0.05 was considered statistically significant. A sample-size analysis was conducted using MedCalc for the reader study. Assuming AUCs of 0.747 and 0.878, correlations of 0.46 and 0.25 in the positive and negative groups, respectively, and using a two-sided $\alpha=0.05$, 146 cases were required to provide 80\% power for the prespecified Round 2 versus Round 3 AUC comparison. The final reader-study dataset included 167 cases. Details are provided in the Supplementary Information. For health economics microsimulation, we used 5,000 paired nonparametric bootstrap replicates to evaluate the 95\% confidence intervals and P-values.

\subsection{Implementation details}
We implemented the project based on the following packages: Python (3.10.12), OpenCV (4.10.0.84), pandas (2.2.2), NumPy (1.26.4), Pillow (10.4.0) and nnUNetv2 (2.6.4). Additionally, the deep learning model is implemented using PyTorch (2.4.0) and Torchvision (0.19.0). Evaluation metrics are calculated using scikit-learn (1.5.1). We conducted the experiments using computational resources from a cluster with eight NVIDIA H100 GPUs.

\backmatter

\bmhead{Data Availability}

The internal and real-world datasets are not currently available for public release, as required by the respective institutional review boards. Requests for access to aggregate data and supporting clinical documents will be considered by an independent review panel based on scientific merit. Data access requests related to this study may be directed to the first author, Haojun Yu (haojunyu@pku.edu.cn). Requests will be reviewed and processed within 6 weeks. The external UK Biobank  and MIMIC-IV datasets are publicly available. Researchers can apply for access through \url{https://www.ukbiobank.ac.uk/} and \url{https://physionet.org/content/mimiciv/3.1/}.

\bmhead{Code Availability}

The code used for the implementation of this study is dependent on internal tooling and infrastructure and is under patent protection (ZL 2026 1 0342889.4 and ZL 2026 1 0303001.6), and thus cannot be publicly released. All experiments and implementation details are described in sufficient detail in the Methods and Supplementary Information (Methods) sections to support replication with non-proprietary libraries. Several major components of our work are available in open-source repositories: nnUNet (\url{https://github.com/MIC-DKFZ/nnUNet}) and ms-swift (\url{https://github.com/modelscope/ms-swift}).

\bmhead{Acknowledgments}

We thank L.C. and R.G. from Qingdao University for helpful suggestions and discussions. We thank L.C., R.G., Z.G., X.H., K.L., X.L., M.S., X. Su., X. Sun., J.W., Q. Wang., C.X., M.Z. and S.Z. for participating in the reader study. H.R. was supported by the National Natural Science Foundation of China (82125026, 82330081) and the ``Medicine Plus'' Joint Research Program of Qingdao University (No. YX2024201). L.W. was supported by the National Natural Science Foundation of China (NSFC92470123, NSFC62276005).

\bmhead{Author contributions}

H.Y., H.R., Y.X., L.W., and A.S. conceived and designed the study. H.R., N.M., H.X., W.Q., B.S. and Z.L. provided the datasets. J.N. and J.P. provided the computing resources. H.Y., J. Zheng., H.X., X.T., Y.G., X.J., W.Y., R.S. and X.Z. carried out data acquisition. H.Y., J. Zheng., Y.G., X.J., Y.Liu., C.L., Z.Z., X.T., X.Z., H.D., W.Y. and R.S. carried out data processing, annotation and quality control. H.Y., Y.X., and C.L. developed the AI models. J. Zheng. developed the platform for the reader study. H.Y., N.M., J. Zheng. and H.R. carried out the reader study. J.X., Y.Lan., and Z.L. carried out the health-economic analyses. H.Y., J. Zheng., H.R., Y.X., G.F., J. Zou., C.L., X.T., Y.G., X.J., Q.Wu., W.Y., R.S., X.Z., J.X., and J.N. wrote and revised the paper. All authors reviewed and approved the final manuscript.

\bmhead{Competing Interests}

The authors declare no competing interests.

\bibliography{sn-bibliography}

\end{document}